\documentclass{article}
\usepackage[preprint]{nips_2018}
\usepackage[utf8]{inputenc} 
\usepackage[T1]{fontenc}    
\usepackage{hyperref}       
\usepackage{url}            
\usepackage{booktabs}       
\usepackage{amsfonts}       
\usepackage{nicefrac}       
\usepackage{microtype}      
\usepackage{graphicx}
\usepackage{tabularx,ragged2e,booktabs,caption}
\usepackage{amssymb}
\usepackage{amsmath}
\usepackage{color}
\usepackage{enumerate}
\usepackage{enumitem}
\newcolumntype{C}[1]{>{\Centering}m{#1}}

\newcolumntype{S}[1]{>{\hsize=.2\hsize}m{#1}}

\title{
Evaluating Artificial Systems for Pairwise Ranking Tasks Sensitive to Individual Differences
}

\author{
  Xing Liu\\
  Graduate School of Information Sciences, Tohoku Univeristy\\
  \texttt{ryu@vision.is.tohoku.ac.jp} \\
  \And
  Takayuki Okatani \\
  Graduate School of Information Sciences, Tohoku University\\
  RIKEN Center for AIP \\
  \texttt{okatani@vision.is.tohoku.ac.jp} \\
  }

\begin{document}
\maketitle

\begin{abstract}
Owing to the advancement of deep learning, artificial systems are now rival to humans in several pattern recognition tasks, such as visual recognition of object categories. However, this is only the case with the tasks for which correct answers exist independent of human perception. There is another type of tasks for which what to predict is human perception itself, in which there are often individual differences. Then, there are no longer single ``correct’’ answers to predict, which makes evaluation of artificial systems difficult. In this paper, focusing on pairwise ranking tasks sensitive to individual differences, we propose an evaluation method. Given a ranking result for multiple item pairs that is generated by an artificial system, our method quantifies the probability that the same ranking result will be generated by humans, and judges if it is distinguishable from human-generated results. We introduce a probabilistic model of human ranking behavior, and present an efficient computation method for the judgment. To estimate model parameters accurately from small-size samples, we present a method that uses confidence scores given by annotators for ranking each item pair. Taking as an example a task of ranking image pairs according to material attributes of objects, we demonstrate how the proposed method works.
\end{abstract}

\section{Introduction}

Standard recognition tasks, in which some entities (e.g., labels) are to be predicted from an input (e.g., an image etc.), can be roughly categorized into two groups:
\begin{itemize}[wide, labelwidth=!, labelindent=0pt] \item [$T_1$:] Tasks in which what to predict is given independent of human perception.
\item [$T_2$:] Tasks in which what to predict is human perception itself.
\end{itemize}
For the sake of explanation, we limit our attention to visual tasks in what follows, although the discussions apply to other modalities. Then, examples of $T_1$ are visual recognition of object categories and individual faces, and examples of $T_2$ are prediction of aesthetic quality (\cite{AVA,aethe2,AADB,aes-survey}) and memorability (\cite{memo}) of images.

In order to build a machine-learning-based system for task group $T_2$, it is first necessary to  materialize what humans perceive from images. This can be performed by, for example, asking human subjects to give a score for an image or asking them to rank multiple images. Then, we consider training artificial systems (e.g., convolutional neural networks) so that they will predict the scores or ranking results as accurately as possible.

As is well recognized, CNNs can now rival humans for several visual recognition tasks (\cite{resnet,googlenet}), when they are properly trained in a supervised manner. Although it may not be widely recognized, this is only the case with task group $T_1$, in which labels to predict are given independent of human perception and thus there should exist correct answers to predict; therefore it is straightforward to define and measure the performance of CNNs. For task group $T_2$, however, it remains unclear whether CNNs can achieve the human level of performance, although the advancement of deep learning arguably has contributed to significant performance boost.

\begin{figure}[!]
\centering
\includegraphics[clip,width=130mm]{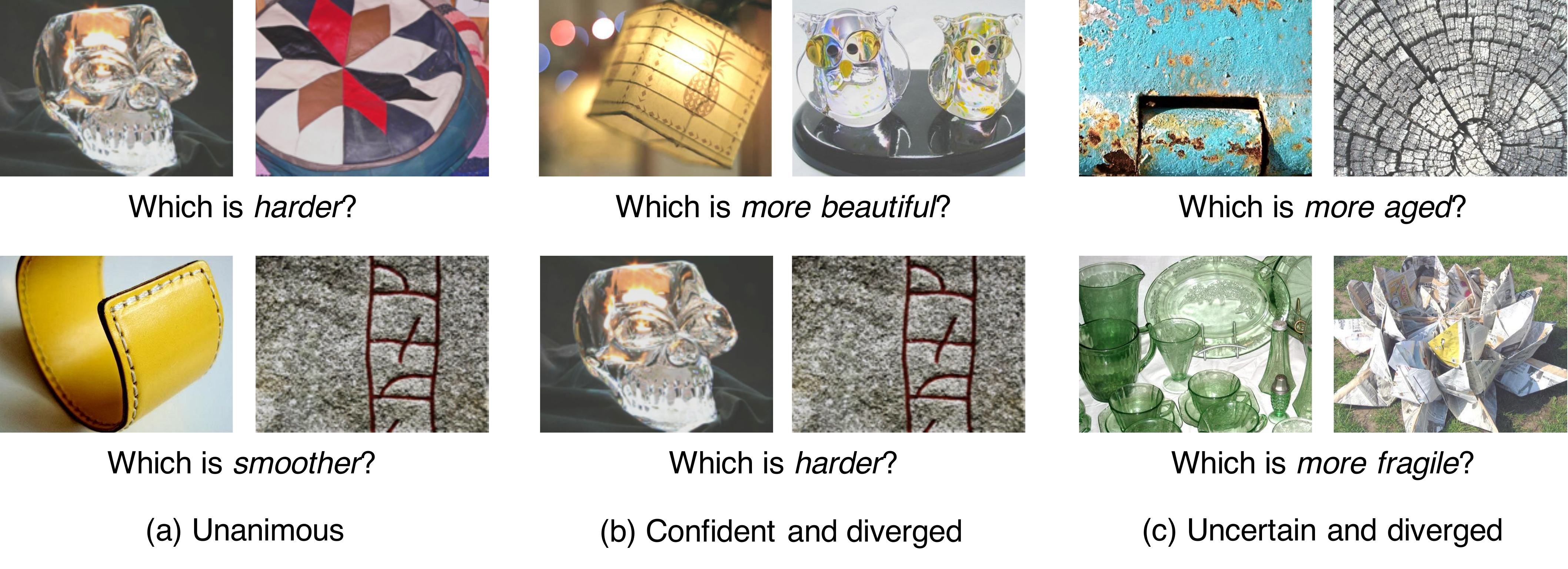}
\caption{
Examples of pairwise ranking of images according to material attributes of objects. Rankings given by different annotators are (a) unanimous, (b) diverged with confidence, or (c) uncertain and diverged. }
\label{iccvface}
\end{figure}

This may be attributable to individual differences of annotators that often emerge in tasks of $T_2$. When there are individual differences, there is no unique  correct answer to predict, which makes it difficult to evaluate the performance of artificial systems.
Figure~\ref{iccvface} shows an example of such cases, that is, pairwise ranking of images according to material attributes of objects. While, for some image pairs and attributes, human annotators will give unanimous ranking as depicted in Figure~\ref{iccvface}(a), for others, they will give diverged rankings. The latter can be divided into two cases, i.e., when the annotators confidently make diverged rankings, which mostly occurs for subjective cases, as in Figure~\ref{iccvface}(b), and when they are uncertain and give diverged rankings, as in  Figure~\ref{iccvface}(c).

In order to build an artificial system that rivals humans for this type of tasks, we need to answer the following questions:
\begin{itemize}[wide, labelwidth=!, labelindent=0pt] 
\item[$Q_1$.] How can we measure its performance or judge its equivalence to humans? 
\item[$Q_2$.] How can we build such an artificial system?
\end{itemize}
In this paper, targeting at pairwise ranking tasks, we attempt to give answers to these questions. 

For $Q_1$, we propose a method that is based on a probabilistic model of ranking results given by human subjects.
Suppose that an artificial system predicts ranking of $N$ image pairs. The proposed method quantifies {\em how probable it is that the same ranking result is generated by human annotators} for the same $N$ image pairs. It properly considers the above individual differences. 

Despite its simplicity, there are a few difficulties to overcome with this approach. One is the difficulty with obtaining an accurate probabilistic model of human ranking from small sample size data\footnote{We have $M$ human annotators rank each of $N$ item pairs. Considering data collection cost, increasing both $M$ and $N$ is prohibitive. As $N$ needs to be large, $M$ has to be small.}. To resolve this, we propose to collect confidence scores from annotators for ranking each item pair, and utilize them to accurately estimate a probabilistic model of human ranking (Sec.\ref{sec:estimate-choice-p}). Thus, our proposal includes an annotation scheme for achieving the goal. Another difficulty is with computational complexity. The presence of individual differences increases the number of ranking patterns that humans can generate for $N$ item pairs, making naive computation infeasible. We present a method that performs the necessary computation efficiently (Sec.\ref{q-value-comp}). Following the proposed data collection scheme, we have created a dataset named Material Attribute Ranking Dataset (MARD), on which we tested the proposed method. The dataset will be made publicly available. 

For $Q_2$, we argue that learning to predict distribution of rankings given by multiple annotators performs better than previous methods. In previous studies (\cite{bestpaper,RRankcnn,ranknet-p}), individual differences are usually ignored; the task is converted to binary classification by taking the majority if there are individual differences. Our experimental results support this argument.  



     

\section{Distinguishing Artificial Systems from Humans}
\label{twomeasures-section}

\subsection{Outline of the Proposed Approach}

We consider a pairwise ranking task using $N$ pairs, where, for instance, two images in each pair $(i=1,2,\ldots,N)$ are to be ranked. We denote a ranking result of a human subject (or an artificial system) by a sequence $ X \triangleq \{ x_{1}, x_{2}, \ldots, x_{N} \}$, where $x_i$ is a binary variable, such that $x_i = 1$ if a subject chooses the first image, and $x_i=0$ if the subject chooses the second image.

Considering the aforementioned individual differences, we introduce a probabilistic model for $X$. Let $p(X)$ be the probability mass function of human generated sequences $X$'s. 
Given a sequence $\mathtt X$ generated by an artificial system, we wish to use $p(X)$ to estimate how probable a human can generate $\mathtt X$. If the probability is very low, it means that $\mathtt X$ is distinguishable from human sequences; then, we may judge the artificial system behaved differently from humans. If it is high, then we cannot distinguish $\mathtt X$ from human sequences, implying than the artificial system behaves similarly to humans, as far as the given task/dataset is concerned.

Ideally, any $X$'s with $p(X)\neq 0$ can be generated from human subjects, and thus it could be possible to use $p(\mathtt X)=0$ or $\neq 0$ for the above judgment. However, this is not appropriate. As the exact $p(X)$ is not available, we have to use an approximate model built upon several assumptions. Moreover, $p(X)$ is estimated from the data collected from human subjects, which could contain noises. It tends to have a long-tail (i.e., many sequences with a very small probability). Thus, the model $p(X)$ may be unreliable particularly for $X$'s with low $p(X)$'s. These sequences are also considered to be minor and eccentric sequences that the majority of humans will not generate. 

Therefore, we use a criterion that excludes such minor sequences with the lowest probabilities. Let $S$ denote a subset of $2^N$ possible sequences for the $N$ pairs. In particular, we consider a subset $S$ with the minimum cardinality $\vert S\vert$ that satisfies the following inequality:
\begin{equation}
  \sum_{X\in S} p(X) > 1-\epsilon,
\end{equation}
where $\epsilon$ is a small number. We denote it by $S_\epsilon$.
$S_\epsilon$ indicates a set of sequences that humans are likely to generate. The complementary set $S_0\setminus S_\epsilon$ contains the above-mentioned minor sequences. 

Given a ranking result $\mathtt X$ created by an artificial system for the same $N$ pairs, we check if ${\mathtt X}$ belongs to $S_\epsilon$, i.e., whether ${\mathtt X} \in S_\epsilon$ or not. We declare ${\mathtt X}$ to be indistinguishable from human results if ${\mathtt   X} \in S_\epsilon$, and to be distinguishable if ${\mathtt X} \notin S_\epsilon$.


\subsection{Model of Human Ranking}
\label{probmodel-for-pairranking}

We now describe the model $p(X)$. 
Assuming that ranking results of different pairs are independent of each other, we model the probability of a sequence $X$ by
\begin{equation}
\label{bimo}
p(X)=p(x_{1}, x_{2}, \ldots, x_{N}) = \prod^{N}_{i=1}p(x_{i}).
\end{equation}
We then model $p(x_i)$ using a Bernoulli distribution with a parameter $\theta_i$, that is, 
\begin{equation}
\label{px}
    p(x_{i}=1) = \theta_i \; \; {\rm and} \;\;\; p(x_{i}=0) = 1-\theta_i.
\end{equation}
Determination of $\theta_i$ will be explained later. Human subjects can provide many different sequences; each sequence $X$ will occur with probability $p(X)$. 

It should be noted that the Bradley-Terry model, a popular model of pairwise ranking, is not fit for our problem. It considers a closed set of items (e.g., sport teams and scientific journals), and is mainly used for the purpose of obtaining a ranking of all the items in the set from observations of ranking of item pairs). On the other hand, in our case, we consider an open set of items. 
Our interest is not with the item set itself but with evaluation of an intelligent system performing the task. 

\subsection{Percentile of a Sequence in Probability-Ordered List }
\label{q-value-comp}

\begin{figure}[!]
\centering
\includegraphics[clip,width=7.5cm]{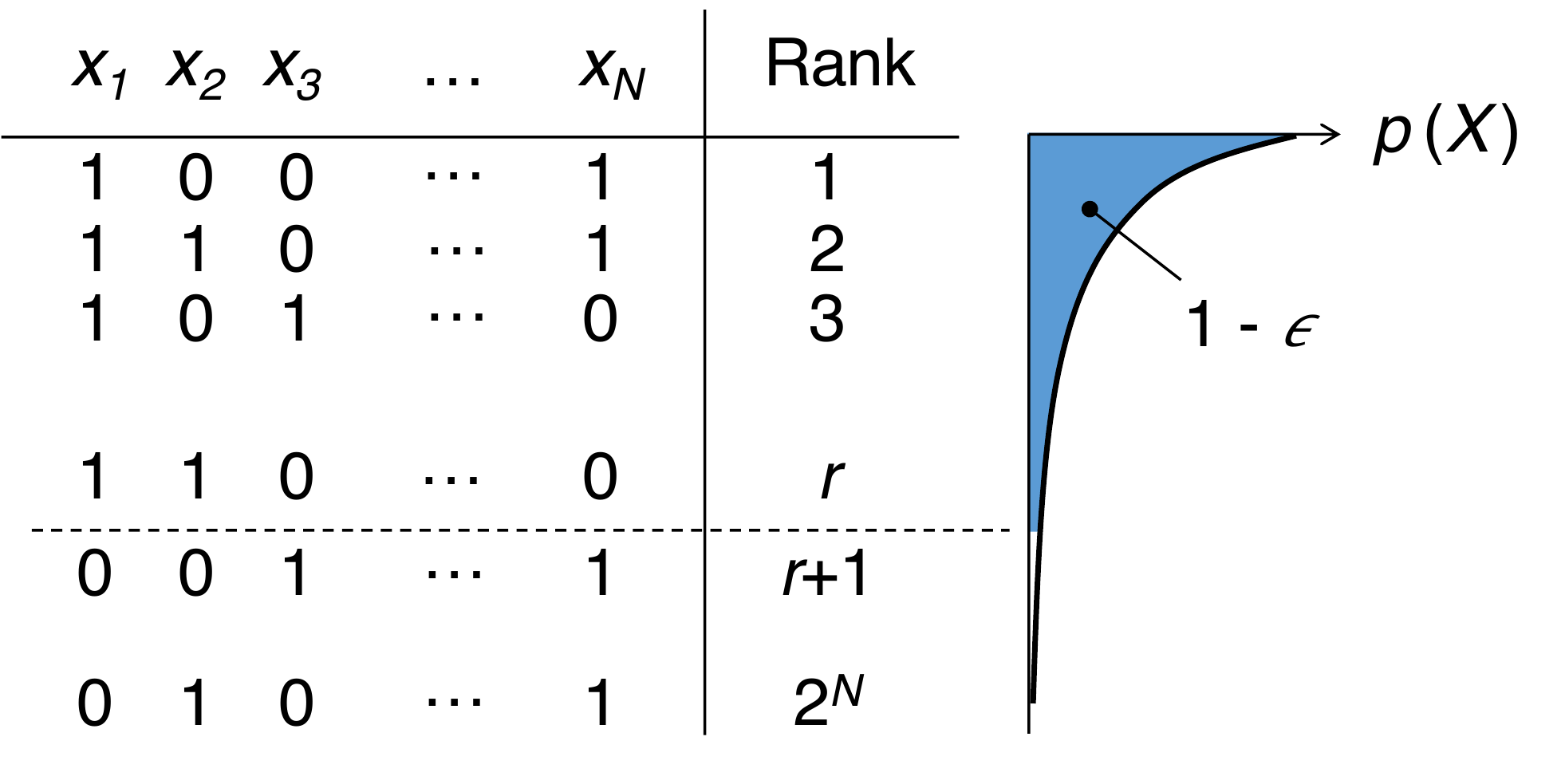}
\caption{
Each $N$ bit sequence represents an instance of pairwise rankings for $N$ item pairs. These sequences are sorted in the descending order of their probabilities. The hatched area of $p(X)$ on the right indicates the cumulative probability of $1-\epsilon$. We check a machine-generated sequence is ranked above the lowest rank $r$. This test is efficiently performed by calculating its percentile value $Q$ and see if $Q<1-\epsilon$. }
\label{human-zone}
\end{figure}

\paragraph{Percentile Value $Q$}
As mentioned above, we want to check whether $\mathtt X \in S_\epsilon$ or $\notin S_\epsilon$. 
To perform this, we calculate the {\em percentile} of $\mathtt{X}$ in the ordered list of all possible sequences $X$'s in the order of $p(X)$. It is defined and computed as follows (see Fig.\ref{human-zone}): {\em i) Sort all (i.e., $2^N$) possible binary sequences $X$'s in descending order of $p(X)$ of (\ref{bimo}) and ii) Compute the percentile of $\mathtt{X}$ (denoted by $Q$) using the cumulative sum  of probabilities from the first rank to the position where $\mathtt{X}$ is ranked. }
Using $Q$ thus computed for the target $\mathtt X$, the condition $\mathtt X \in S_\epsilon$ is equivalent to $Q\leq 1-\epsilon$ (and $\mathtt X \notin S_\epsilon$ is equivalent to $Q > 1-\epsilon$).
It is noteworthy that $Q$ represents how close the sequence is to 
the most probable ranking result of humans, which corresponds to $Q=0$.

\paragraph{Efficient Computation of $Q$}
When $N$ is large, it is not feasible to naively perform the above procedure, as the number of possible sequences explodes. It is also noted that 
$\theta_i$ may differ for each $i$, and thus the standard statistics of binomial distribution cannot be computed for the whole sequence. 
Thus, we group the elements having the same $\theta_i$ to a subsequence, which are specified by an index set $I_g\triangleq\{i\,\vert\,\theta_i=\theta_g\}$ for a constant $\theta_g$.
Suppose that this grouping splits $X$ into $G$ subsequences $X_1,X_2, \ldots,X_G$. Using the independence of the elements, we have
\begin{equation}
    p(X)=\prod_{g=1}^G p(X_g).
\end{equation}
In this grouping, we may redefine the variable $x_i$ by swapping the first and second images so that $\theta_i\geq 0.5$.
This enables us to minimize the number $G$ of groups without loss of generality to improve computational efficiency as will be discussed below.

Let $n_g$ be the number of elements belonging to $X_g$ (i.e., $n_g=\lvert I_g\rvert$), and $\theta_g$ be their Bernoulli parameter (i.e. ${\theta_g=\theta_i}$ for any $i\in I_g$). Then, the probability of $X_g$ being a sequence $\mathtt{X}_g$ is computed by
\begin{equation}
\label{xg}
 P(X_g=\mathtt{X}_g) = \theta_g^{k_g}{(1 - \theta_g)}^{(n_g - k_g)},
\end{equation}
where $k_g$ is the number of 1's (i.e., $x_i=1$) in $\mathtt{X}_g$. 
Note that the number of possible sequences having $k_g$ 1's is $\binom{k_g}{n_g}$, and each of them has the same probability computed as above. 
For the entire sequence, the probability of $X$ being a sequence $\mathtt{X}$ is given by
\begin{equation}
    P(X=\mathtt{X}) =
    \prod_{g=1}^G P(X_g=\mathtt{X}_g).
    \label{eq:seq1}
\end{equation}
Using \eqref{eq:seq1} along with \eqref{xg}, we can compute the probability that a sequence $\mathtt{X}$ consists of subsequences $\mathtt{X}_g  ({g=1,2,\ldots,G}$), each of which has $k_g$ 1's. The number of such sequences is calculated by $M(k_1,k_2,\ldots,k_g)\triangleq\prod \limits_{g=1}^G \binom{k_g}{n_g}$, and each sequence has the same probability. 

Now, we consider sorting all $2^N$ sequences of $X$. We construct a sequence $\{k_1,k_2,\ldots,k_G \}$ by choosing each of its elements $k_g$ (the number of 1's in the $g^{th}$ group) from $[0,n_g]$. We obtain $M(k_1,k_2,\ldots,k_G)$ sequences having the same probability by employing this assignment scheme. We denote the $j^{th}$ assignment by $K_j$ ($j=1,2,\ldots,J)$, where $J$ is the number of possible assignments, which is given by $\prod_{g=1}^G (n_g+1)$. (Note that $\sum_{j=1}^J M(K_j)=2^N$.) 

As it is not necessary to sort sequences having the same probability, we only need to sort $J$ blocks of sequences instead of $2^N$ individual sequences. For each  $j^{th}$ block associated with an assignment $K_j$, we use \eqref{eq:seq1} for computation of the probability of each sequence belonging to this block. Let $P_j$ be this probability. Then, using $P_j$, we sort $J$ blocks, which can be done much more efficiently than sorting $2^N$ sequences. 

Finally, we compute $Q$ for a sequence $\mathtt{X}$.
In order to compute its rank, we count the number of 1's in $\mathtt{X}$ (more specifically, the number of 1's in each $\mathtt{X}_g$ of $\mathtt{X}=\{ \mathtt{X}_1,\mathtt{X}_2,\ldots,\mathtt{X}_G \}$, and find the block $j_\mathtt{X}$ to which $\mathtt{X}$ belongs. Let $I_\mathtt{X}$ be the index set of blocks ranked higher than $j_\mathtt{X}$, i.e., $I_\mathtt{X}\triangleq\{j\,\vert\,P_j\geq P_{j_\mathtt{X}}\}$. Then, the cumulative probability down to block $j_\mathtt{X}$ (including it) can be computed by
\begin{equation}
 Q = \sum_{j\in I_\mathtt{X}} M(K_j)P_j.
\end{equation}

\label{sec:qk}

\section{Estimation of the Bernoulli Parameter Using Confidence Scores}
\label{sec:estimate-choice-p}

\subsection{Small-sample Estimate of $\theta_i$}

We modeled human ranking by the Bernoulli distribution as in (\ref{px}). We now consider how to estimate its parameter $\theta_i, \forall i$. Suppose $n_i$ subjects participate in ranking the $i^{th}$ image pair, $\forall i$. Let $n_i'(\leq n_i)$ be the number of subjects who chose the first image. Considering a pairwise ranking task with exclusive choice, the number of subjects who chose the second image is ${n_i-n_i'}$. Then, the maximum likelihood estimate (MLE) of $\theta_i$ is immediately given by
\begin{equation}
\label{pxest}
    \theta_i = \frac{n_i'}{n_i}, \forall i.
\end{equation}

Despite its simplicity, this method could have an issue when the subjects unanimously choose the same image of an image pair, i.e., either $n_i'=n_i$ or $n_i'=0$. In this case, the above MLE gives $\theta_i=1$ or $\theta_i=0$, which leads to $p(x_i=0)=0$ or $p(x_i=1)=0$. However, this result is quite sensitive to (in)accuracy of $\theta_i$ and thus results may not be useful. If a CNN chooses the one with $p(x_i)=0$ for only a single unanimous pair, then $p(X)$ immediately vanishes irrespective of ranking of other pairs (i.e., $Q=100$\%), declaring that this CNN behaves completely differently from human. The estimate (\ref{pxest}) could be inaccurate if $n_i$ is not large enough. Although this issue will be mitigated by using a large $n_i$, it will first increase the cost of data collection; it will also increase the computational complexity of $Q$ (because the number $G$ of groups having an identical $\theta_i$ tends to increase).

\subsection{A More Accurate Estimate Using a Confidence Score}

Therefore, we instead consider collecting additional information from human subjects. For each image pair $i$, we ask them to additionally give a {\em confidence score} of their ranking. We use a score $s\in \{0, 1, 2\}$, which correspond to ``not confident'', ``somewhat confident'' and ``very confident'', respectively. As we ask $n_i$ subjects for a single image pair $i$, we introduce an index $h(=1,\ldots,n_i)$ to represent each subject and denote the ranking choice and score of $h$-th subject by $x_{ih}$ and $s_{ih}$, respectively. Let $X^{(i)}=[x_{i1},\ldots,x_{in_i}]$ and $S^{(i)}=[s_{i1},\ldots,s_{in_i}]$. 
Assuming independence of individual annotations, we have
\begin{equation}
\label{eq:estimate-1}
  \begin{aligned}
     p(X^{(i)},S^{(i)}|\theta_i) &= \prod_{h=1}^{n_i}p(x_{ih}, s_{ih} |\theta_i) \\
                          &=\prod_{h=1}^{n_i}p(s_{ih}|x_{ih}, \theta_i)p(x_{ih}|\theta_i).
  \end{aligned}
\end{equation}

We use this model to perform maximum likelihood estimation for the unanimously ranked pairs. 
Suppose that all $n_i$ subjects chose the first image. Then we have
\begin{equation}
  \prod_{h=1}^{n_i}p(x_{ih}=1|\theta_i) = {\theta_i}^{n_i}.
\end{equation}
To model $p(s_{ih}|x_{ih},\theta_i)$, we consider the probability of occurrence of each score $s=0,1,2$; we denote it by $q_s^{(i)} \equiv p(s_{ih}=s)$. Then we have
\begin{equation}
  \prod_{h=1}^{n_i}p(s_{ih}|x_{ih}=1, \theta_i) = {q_0}^{n_0}{q_1}^{n_1}{q_2}^{n_2},
\end{equation}
where $n_0$, $n_1$ and $n_2$ denotes the number of subjects who chose confidence scores 0, 1, and 2, respectively; we omit the superscript in $q_0$, $q_1$, $q_2$ etc. for simplicity. Thus, from \eqref{eq:estimate-1} we have
\begin{equation}
\label{eq:estimate-f}
  p(X^{(i)}={\mathbf 1}, S^{(i)}|\theta_i) = {\theta_i}^{n_i}{q_0}^{n_0}{q_1}^{n_1}{q_2}^{n_2}.
\end{equation}

Given the observations $S^{(i)}$ and $(n_0,n_1,n_2)$, we want to maximize the likelihood \eqref{eq:estimate-f} with respect to $\theta_i$ as well as the introduced unknowns $q_0$, $q_1$, and $q_2$. As they are probabilities, there are a few constraints for $q_0$, $q_1$, and $q_2$ defined as 
\begin{subequations}
\label{eq:qconsts}
\begin{align}
    & q_0+q_1+q_2=1,\\
    & 0\leq q_0,q_1,q_2 \leq 1.
\end{align} 
\end{subequations}
We then assume a relation between the confidence scores and $\theta_i$ leading to the following equation:
\begin{equation}
\label{eq:estimate-important_constraint}
\frac{1}{2}q_0 + \frac{3}{4}q_1 + q_2 = \theta_i.
\end{equation}
This equation indicates that the three scores, {\em not confident}, {\em confident}, and {\em very confident}, are mapped to $\theta_i=0.5$, 0.75, and 1.0, respectively. In other words, subjects who are not confident irrespective of the choice $x_i=1$ will choose the other with 50\% in the future; those who are very confident will always do the same choice in the future; those who have intermediate confidence will do the same with the intermediate probability 75\% in the future. 

Using the constraints (\ref{eq:qconsts}) and (\ref{eq:estimate-important_constraint}), we can maximize the likelihood (\ref{eq:estimate-f}) with respect to the unknowns. 
To be specific, we eliminate, say, $q_0$ and $q_1$, from (\ref{eq:estimate-f}) using (\ref{eq:qconsts}a) and (\ref{eq:estimate-important_constraint})  and maximize it for $\theta_i$ and $q_2$ with the inequality constraints (\ref{eq:qconsts}b). Any numerical constrained maximization method can be used for this optimization. As a result, we have the MLE for $\theta_i$. It should be noted that we use this estimation method only when the ranking results are unanimous; we use the standard estimate (\ref{pxest}) otherwise.

\paragraph{Discussion on the Use of Confidence Score}
As is described above, we propose to use a confidence score to estimate each $\theta_i$. An alternative use of confidence scores is to simply eliminate the ranking results with a confidence score less than 2 and go with only maximally confident ranking results. We do not adopt this approach due to the following reason. We think that there are two cases for how individual differences emerge (see also Fig.\ref{iccvface}): i) annotators make selections confidently, which are nevertheless split (subjectivity); and ii) annotators make selections without confidence, which makes their decisions fluctuated and then split (uncertainty). Elimination of samples with score 0 or 1 will remove data of type (ii). This will be fine when we are only interested in data of type (i), which seems to be the case with most existing studies. On the other hand, our study also considers data type (ii). Suppose, for instance, the case where people perceive very similar glossiness for two different objects, e.g., a plastic cup and a glass mug; uncertainty will be fairly informative for such cases. This is why we attempt to model the uncertainty instead of eliminating it. As we are interested in modeling individual differences, we don’t eliminate samples with low inter-rater agreement, either. 

\section{Experimental Results}

\subsection{Material Attribute Ranking Dataset}
\label{MARD-section}

To test our method, we have created a dataset of a pairwise ranking task. It has been shown (\cite{mate_attr_flem,mate_attr_flem2}) that humans can visually perceive fairly accurately the ``material attribute'' of an object surface, such as hardness, coldness, lightness etc. 
Motivated by this, we chose a task of ranking a pair of images according to such material attributes. We consider  thirteen material attributes, namely {\em aged, beautiful, clean, cold, fragile, glossy, hard, light, resilient, smooth, sticky, transparent}, and {\em wet}. 


The dataset, which we name Material Attribute Ranking Dataset (MARD), consists of 1,000 training and 300 testing samples for each of the thirteen attributes. Each sample contains ranking results of an image pair of five Mechanical Turkers. For test samples for which the initial five Turkers make a unanimous selection, the dataset provides additional ranking results of ten more Turkers as well as three-level scores of their confidence in their ranking. 

It should be noted that MARD differs from any of similar existing datasets of pairwise ranking, e.g., Emotion Dataset (\cite{emotion_dataset}). The existing datasets simply discard individual differences by taking the majority of nonunanimous annotations, which is equivalent to regarding the most probable ranking result of humans as the only correct prediction. Although this greatly simplifies the problem, this makes it impossible to consider individual differences in any ways. 

\paragraph{Details of Creation of MARD}  MARD uses images of the Flicker Material Database (FMD) (\cite{FMD-cite}), a benchmark dataset of classification of ten materials categories. We split 1,000 FMD images into two sets of non-overlapping 500 images (one set is used for training and the other set is used for testing). We then created 1,000 and 300 image pairs by randomly choosing images from each set, respectively. We asked five Turkers to rank each image pair in terms of each of the thirteen material attributes by showing the paired images in a row. (The same image pairs were used for all the attributes.) Specifically, we asked them to choose one of three options, the first image, the second image, and ``unable to decide'', for each pair. We discarded the image pairs with three or more ``unable to decide'' in the training set and those with one or more ``unable to decide'' in the test set. 
For the image pairs of the test set which were ranked unanimously for an attribute by the five Turkers, we further have ten more Turkers rank the same image pair and attribute. In this second task, we removed the ``unable to decide'' option, and also asked the Turkers to additionally provide their confidence on their ranking by choosing one of three level confidence, i.e., ``Not confident'', ``Somewhat confident'', and ``Very confident''.
The confidence scores are used in order to estimate the parameter $\theta_i$ to mitigate the sensitivity issue with unanimous ranking, as was discussed in Section \ref{sec:estimate-choice-p}.




\subsection{Evaluation of Differently Trained CNNs}
\label{sec:exp_mard}

Using the MARD, we conducted experiments to test our evaluation method. To see how it evaluates different prediction methods, we consider four methods. The first three are existing methods that convert the task into binary classification by regarding the majority of human ranking results as the correct label to predict. For the fourth method, we present a method that considers the individual differences as they are. 
\begin{itemize}
    \item RankCNN (\cite{RRankcnn}): A CNN is trained to predict binary labels by minimizing weighted squared distances computed for each training sample to its binary label. 
    \item RankNet$_h$ (\cite{bestpaper}): In their original work, a linear classifier is trained to predict binary labels by searching for maximum-margin hyper-planes in a feature space. In this study, we instead employ a hinge-loss.
    \item RankNet$_p$ (\cite{ranknet-p}): A neural network (a CNN in this study) is trained to predict binary labels by minimizing the cross-entropy loss.
    \item RankDist (ours): A CNN is trained to predict the distribution of ranking of each item pair. To be specific, it is trained to predict the ML estimate (\ref{pxest}) of the Bernoulli parameter $\theta_i$. (Recall that training samples do not provide scores.) The  cross-entropy loss is used. 
\end{itemize}



We applied the above four methods to the training set of the MARD. 
For all the four methods, we use the same CNN, VGG-19 (\cite{vgg}). 
It is first pretrained on the ImageNet and fine-tuned on the EFMD (\cite{yanzhang}). Then it is trained using the MARD where its ten lower weight layers are fixed and the subsequent layers are updated. It is used in a Siamese fashion, providing two outputs, from which each loss is computed. The losses for all the thirteen adjectives are summed and minimized. Note that the four methods differ only in their employed losses.

Table \ref{tab:mard_q-value-vgg19} shows the $Q$ values of the four methods for the thirteen attributes that are computed using the test set of the MARD. When choosing $90.0\%$ for the threshold for $Q$, any ranking results with $Q\geq 90.0\%$ are declared to be distinguishable from human ranking. It is observed that the number of such attributes is 3, 2, 3, and 1, for RankCNN, RankNet$_h$, RankNet$_p$, and RankDist, respectively. RankDist tends to provide smaller $Q$'s for many attributes, indicating that its ranking results are closer to the most probable human results than other three. The three methods show more or less similar behaviours. 

In summary, the proposed evaluation method enables to show which method provides ranking that are (in)distinguishable from human ranking for which attribute. For instance, the ranking results for the attribute {\em hard} are the most dissimilar to human ranking, implying that there is room for improvements. Our method also makes it possible to visualize the different behaviours of the four methods; in particular, RankDist that considers individual differences performs differently from the three existing methods that neglect individual differences.



\begin{table}
\centering
\caption{Evaluation of four differently trained CNNs on the MARD dataset. Numbers which are larger than $90\%$ are displayed in bold fonts, indicating that the ranking results are distinguishable from human-generated results. } 
\label{tab:mard_q-value-vgg19}
\begin{minipage}{0.6\linewidth}
\begin{tabularx}{\linewidth}{@{}C{0.5in}C{0.5in}C{0.5in}C{0.5in}C{0.5in}@{}}\toprule[1.5pt]
  \small(VGG19)&\small RankCNN & \small RankNet$_{h}$& \small RankNet$_{p}$& \small RankDist\\\bottomrule[0.5pt]

  aged& 0.7& 1.9& 4.0& 3.4\\
  beautiful& 39.8& 38.9& 32.3& 38.3\\
  clean& 31.6& 5.7& 27.9& 16.3\\
  cold& 1.8& 8.4& 10.7& 0.2\\
  fragile& {\bf 99.3}& {\bf 97.4}& {\bf 98.7}& 89.1\\
  glossy& 72.4& 88.0& {\bf 90.9}& 32.9\\
  hard& {\bf 100}& {\bf 100}& {\bf 100}& {\bf 93.8}\\
  light& 35.9& 20.8& 36.5& 4.9\\
  resilient& 18.6& 26.9& 16.5& 4.3\\
  smooth& 54.3& 61.3& 71.5& 33.1\\
  sticky& 31.2& 31.2& 10.0& 12.5\\
  transparent& {\bf 100}& 10.0& 7.4& 8.7\\
  wet& 25.3& 27.2& 7.9& 0.3\\
\bottomrule[1.5pt]
\end{tabularx}
\end{minipage}
\end{table}

\section{Summary}

In this study, we have discussed how to compare artificial systems with humans for the task of ranking a pair of items. We have proposed a method for judging if an artificial system is distinguishable from humans for ranking of $N$ item pairs. More rigorously, we check if an $N$-pair ranking result given by an artificial system is distinguishable from those given by humans. It relies on a probabilistic model of human ranking that is based on the Bernoulli distribution. We have proposed to collect confidence scores of ranking each item pair from annotators and utilize them to estimate the Bernoulli parameter accurately for each item pair. We have also shown an efficient method for the judgment that calculates and uses the percentile value $Q$ of the target $N$-pair ranking result. Taking annotation noises and inaccuracies with the models into account, $Q$ is compared with a specified threshold (e.g., 90.0\%); if it is smaller than the threshold, we declare the artificial system is indistinguishable from humans for rankings of the $N$-item pairs. The value $Q$ may also be used as a measure of how close the ranking result of the artificial system is to the most probable ranking result of humans.




\bibliography{nips2018}

\end{document}